\documentclass[10pt,twocolumn,letterpaper]{article}

\usepackage[pagenumbers]{cvpr} 

\usepackage[dvipsnames]{xcolor}

\definecolor{cvprblue}{rgb}{0.21,0.49,0.74}
\usepackage[pagebackref,breaklinks,colorlinks,citecolor=cvprblue]{hyperref}

\def\confName{CVPR}
\def\confYear{2024}

\title{DreamControl: Control-Based Text-to-3D Generation with 3D Self-Prior}

\author{%
  Tianyu Huang$^{1,3}$ \ Yihan Zeng$^{2}$ \ Zhilu Zhang$^{1}$ \ Wan Xu$^{1}$ \ Hang Xu$^{2}$ \\ Songcen Xu$^{2}$ \ Rynson W. H. Lau$^{3}$ \ Wangmeng Zuo$^{1}$\footnotemark[2] \\
  \textsuperscript{1}Harbin Institute of Technology \ \textsuperscript{2}Huawei Noah's Ark Lab \ \textsuperscript{3}City University of Hong Kong \\
}

\usepackage{amsmath,amsfonts,bm}
\usepackage{algorithm}
\usepackage{algorithmic}

\begin{document}
\maketitle
\begin{abstract}
3D generation has raised great attention in recent years. With the success of text-to-image diffusion models, the 2D-lifting technique becomes a promising route to controllable 3D generation. However, these methods tend to present inconsistent geometry, which is also known as the Janus problem. We observe that the problem is caused mainly by two aspects, \textit{i.e.}, viewpoint bias in 2D diffusion models and overfitting of the optimization objective. To address it, we propose a two-stage 2D-lifting framework, namely DreamControl, which optimizes coarse NeRF scenes as 3D self-prior and then generates fine-grained objects with control-based score distillation. Specifically, adaptive viewpoint sampling and boundary integrity metric are proposed to ensure the consistency of generated priors. The priors are then regarded as input conditions to maintain reasonable geometries, in which conditional LoRA and weighted score are further proposed to optimize detailed textures. DreamControl can generate high-quality 3D content in terms of both geometry consistency and texture fidelity. Moreover, our control-based optimization guidance is applicable to more downstream tasks, including user-guided generation and 3D animation. The project page is available at \url{https://github.com/tyhuang0428/DreamControl}.
\end{abstract}    
\section{Introduction}
Digital 3D content plays a crucial role in various fields, including medicine, education, entertainment, \textit{etc}. Generating 3D content in an automatic system is thus raising more and more attention. 
Recently, with the success of score distillation sampling~\cite{poole2022dreamfusion} (SDS), 2D-lifting technique~\cite{poole2022dreamfusion,lin2023magic3d,metzer2023latent,Chen_2023_ICCV,wang2023prolificdreamer} becomes a promising route to 3D generation, where the pre-trained 2D diffusion models~\cite{nichol2022glide,saharia2022photorealistic,rombach2022high} are utilized to optimize 3D representations. 
Compared with methods~\cite{nichol2022point,cheng2023sdfusion,wei2023taps3d,jun2023shap,huang2023textfield3d} that are supervised with 3D assets, 2D-lifting technique is capable of generating high-fidelity 3D textures in open-world scenarios.

\begin{figure}[t]
    \centering
    \includegraphics[width=0.47\textwidth]{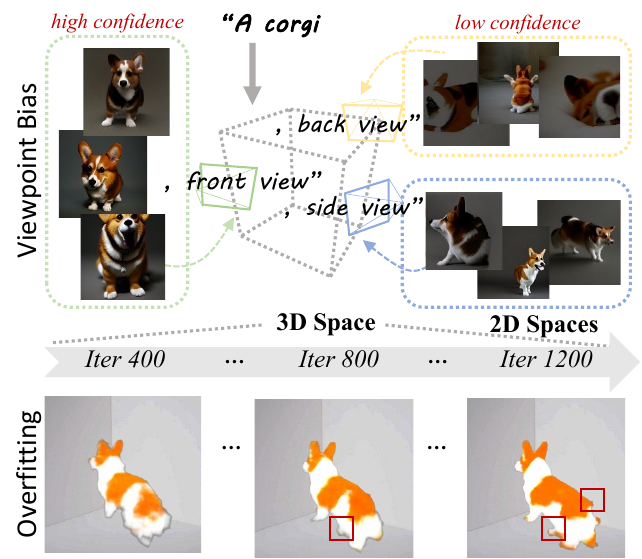}
    \caption{Main causes of inconsistent 3D generation. Images sampled by 2D diffusion models are biased in viewpoint distribution. The generation confidence decreases as the viewpoint turns from front to back. 3D representations are thus gradually overfitted to the highest probability image during the optimization, generating artifacts as shown in red b-boxes.}
    \label{fig:teaser}
\end{figure}

Nonetheless, the results generated by 2D-lifting methods tend to present inconsistent 3D geometry, \textit{e.g.}, multi-face, also known as the Janus problem. 
Some recent works~\cite{liu2023zero,zhao2023efficientdreamer,shi2023mvdream,liu2023syncdreamer,long2023wonder3d,li2023sweetdreamer} attribute this problem primarily to the lack of view awareness in 2D diffusion models. 
They propose to incorporate 3D prior knowledge into 2D diffusion models, thus learning to perceive view-dependent conditions. 
Albeit the improvement of geometry consistency, it compromises texture fidelity and fine-grained details, as the used 3D contents are mostly created manually in a cartoonish style.
Moreover, the limited scale of available 3D assets affects the generalizability of these methods, making it challenging to acquire a comprehensive 3D prior.

\begin{figure*}[t]
    \centering
    \includegraphics[width=\textwidth]{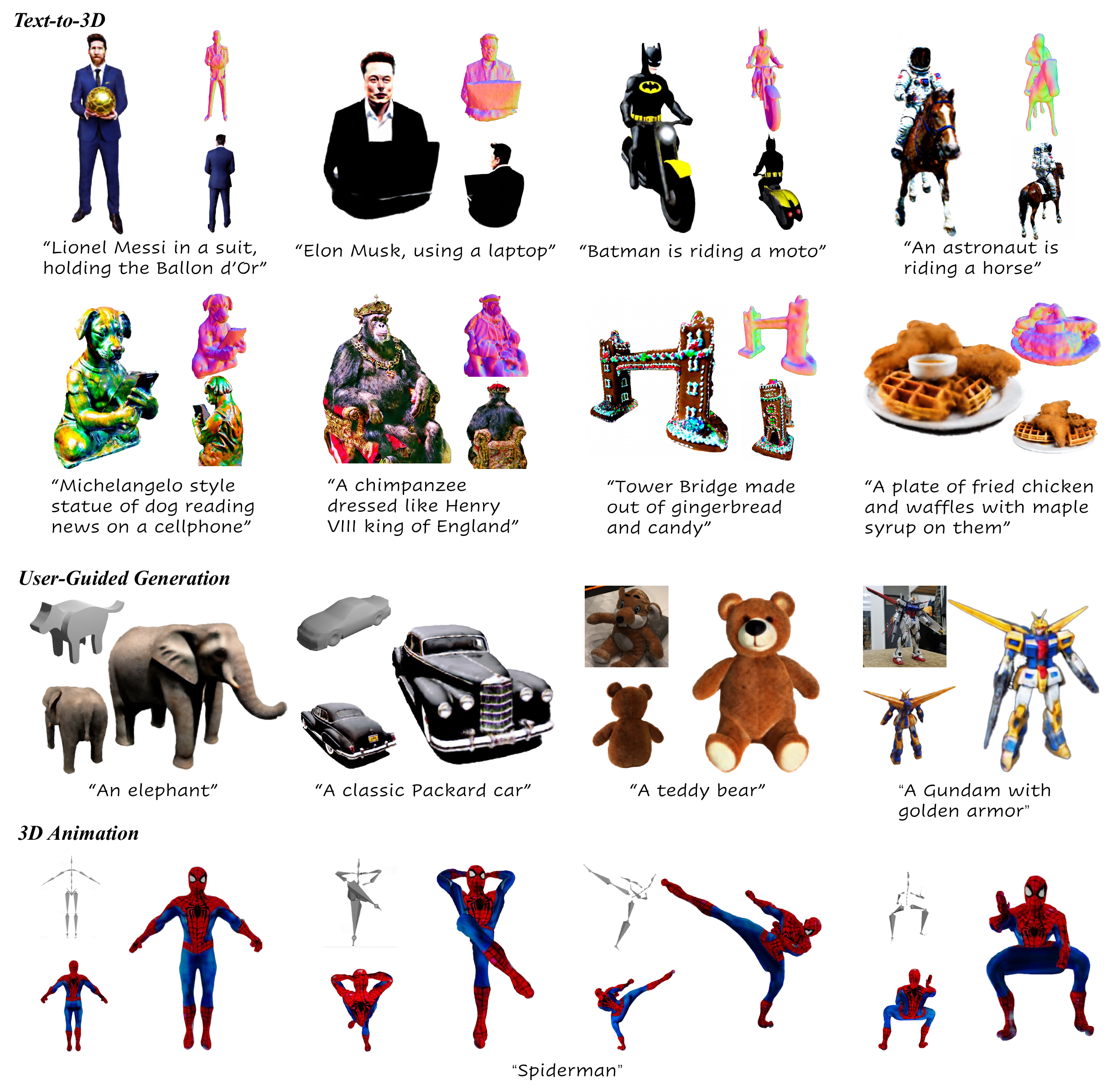}
    \caption{DreamControl can generate diverse 3D content with high-consistency geometries and high-fidelity textures. Beyond text-to-3D generation, our control-based guidance is applicable to controllable generation tasks, including user-guided generation and 3D animation.}
\end{figure*}

Rethinking the optimization target of SDS, \textit{i.e.}, \textit{creating 3D models that look like good images when rendered from random angles}, we observe that the causes of 3D inconsistency can be further divided into two terms: (1) viewpoint bias in 2D diffusion models; (2) overfitting of the optimization objective.
Take neural radiance fields~\cite{mildenhall2021nerf} (NeRF) in SDS as an example:
Points in NeRF scenes are supervised by the casting rays from uniformly sampled viewpoints, while the viewpoint distribution of 2D diffusion models is biased, as shown in Figure~\ref{fig:teaser}. 
Under the optimization of SDS, all the rendered images may overfit to the highest-probability image generated by the diffusion model. 
In other words, all the views of the 3D model look similar to one specific image, giving rise to the Janus problem.

Based on this observation, we aim to leverage the 3D representation before overfitting as a self-generated 3D prior, namely 3D self-prior. Accordingly, we propose DreamControl, a two-stage 2D-lifting framework that maintains self-priors by control-based distillation. 
Specifically, we optimize coarse NeRF scenes as 3D self-prior and then generate fine-grained objects with prior-based control. 
In the first stage, we adopt SDS to construct a coarse shape that keeps good geometry consistency. 
For alleviating possible 3D artifacts, an adaptive viewpoint sampling is proposed to adjust the viewpoint distribution of diffusion models, and a boundary integrity metric is proposed to avoid the overfitting of optimization. 
In the second stage, we regard the 3D prior as a conditional input and deploy ControlNet~\cite{zhang2023adding} to supervise the generation, thus obtaining a detailed texture while maintaining the geometry of the prior.
Considering the diversity of ControlNet can be easily constrained by fixed conditions, we propose control-based score distillation, in which a conditional LoRA and a weighted score are presented to stabilize the optimization process. 

Extensive experiments on text-to-3D generation demonstrate that DreamControl can obtain high-quality 3D content regarding geometry consistency and texture fidelity. 
Benefiting from the control-based guidance, our framework can be further applied to more downstream tasks, including user-guided generation and 3D animation.

Our contributions can be summarized as:
\begin{itemize}
    \item We propose to optimize NeRF as 3D self-prior, where adaptive viewpoint sampling and boundary integrity metric are suggested to alleviate inconsistent generation.
    \item We propose a control-based score distillation to maintain geometries in self-prior, where conditional LoRA and weighted score are presented to stabilize the optimization.
    \item Our two-stage framework, namely DreamControl, can generate high-quality 3D content in text-to-3D generation, and the control-based guidance can be further applicable to more downstream tasks.
\end{itemize}
\section{Related Work}
\noindent\textbf{Text-to-3D Generation.}
Text-to-3D generation has witnessed rapid progress in recent years, in which methods can generally be split into two categories, \textit{i.e.}, 3D supervised and 2D lifting. 3D supervised methods~\cite{nichol2022point,cheng2023sdfusion,wei2023taps3d,jun2023shap,yu2023towards,huang2023textfield3d} train generators with text-3D data. Albeit the efficiency to generate solid 3D content, these methods lack generalizability due to the limited scale of available 3D data. In contrast, 2D lifting methods~\cite{poole2022dreamfusion,lin2023magic3d,metzer2023latent,Chen_2023_ICCV,wang2023prolificdreamer} take advantage of 2D diffusion models, distilling 3D representations to 2D priors. Although presenting photorealistic generation, these methods can easily fall into 3D inconsistency issues, also known as the famous Janus problem. To address the problem, recent works~\cite{liu2023zero,zhao2023efficientdreamer,shi2023mvdream,liu2023syncdreamer,long2023wonder3d,li2023sweetdreamer} attempt to incorporate 3D prior into 2D diffusion models. Since the prior is trained with limited 3D data, these methods still suffer from the lack of generalizability. Moreover, their generation may lose high-fidelity texture due to the cartoonish style of 3D data. In this work, we propose to optimize a coarse NeRF representation as a training-free 3D prior, enhancing generation consistency while keeping texture fidelity.

\noindent\textbf{Controllable Generation.} Text input is a flexible control in 3D tasks~\cite{huang2023clip2point}, while other conditions like image~\cite{ruiz2023dreambooth,wei2023elite}, video~\cite{zhang2023controlvideo,chen2023control}, and 3D sketch~\cite{metzer2023latent} are also available for guiding generation. ControlNet~\cite{zhang2023adding} supports text-to-image synthesis with additional conditions, \textit{e.g.}, edge, normal, \textit{etc}. It allows 3D generators to create 3D content with the guidance of 2D sketch~\cite{chen2023control3d}, depth~\cite{yu2023hifi}, and even video~\cite{shao2023control4d}. In this work, we use 2D conditions to maintain geometry consistency, which are rendered from our 3D prior.

\section{Preliminaries}
\noindent\textbf{NeRF}~\cite{mildenhall2021nerf} (Neural Radiance Fields) is a widely-used 3D representation, which combines neural networks with graphical principles. A multilayer perceptron (MLP) $\theta$ is trained to predict the color $\textit{RGB}$ and density $\bm{\sigma}$ of sampling points in the 3D space, supervised by the total squared error between the rendered and ground-truth pixel colors.
Given a camera pose $c$, images $x$ are rendered by the density summation of colored points. We represent the rendering process as $x = \mathbf{g}(\theta, c)$, in which $\mathbf{g}$ denotes the renderer.

\noindent\textbf{SDS}~\cite{poole2022dreamfusion} (Score Distillation Sampling) distills the parameters of 3D representation NeRF ($\theta$) to a pre-trained 2D diffusion models ($\phi$). Given a text prompt $y$, it optimizes the rendered image $x_t$ with the predicted noise in the timestep $t$. The gradient can be formulated as,
\begin{equation}
\label{eq:sds}
\nabla_{\theta}\mathcal{L}_{\text{SDS}}(\theta) = 
\mathbb{E}_{t,\bm{\epsilon}}\left[
    \omega(t)
    \left( \hat{\bm{\epsilon}}_{\phi}(\bm{x}_t,t,y) - \bm{\epsilon} \right)
    \frac{\partial \bm{x}}{\partial \theta}
\right],
\end{equation}
where $\hat{\bm{\epsilon}}_{\phi}$ is the noise predicted by $\phi$.

\noindent\textbf{VSD}~\cite{wang2023prolificdreamer} (Variational Score Distillation) supposes the corresponding 3D scene given a textual prompt as a distribution range, rather than a single point as in SDS, significantly improving quality and diversity of 3D generation. It proposes a particle-based update strategy via the Wasserstein gradient flow to optimize a 3D distribution,
\begin{equation}\label{eq:vsd}
\mathbb{E}_{t,\bm{\bm{\epsilon}}}\left[
    \omega(t)
    \left( \hat{\bm{\epsilon}}_{\phi}(\bm{x}_t,t,y) - \hat{\bm{\epsilon}}_{\theta}(\bm{x}_t,t,c,y)  \right)
    \frac{\partial \bm{x}}{\partial \theta}
\right],
\end{equation}
where $\hat{\bm{\epsilon}}_{\theta}(\bm{x}_t,t,c,y)$ is the noise predicted by rendered images. In practice, it can be regarded as LoRA~\cite{hu2021lora} (Low-Rank Adaption) conditioned with camera poses $c$, which is supervised by a standard diffusion loss as,
\begin{equation}\label{eq:lora}
\mathbb{E}_{t\sim\mathcal{U}(0,1),\bm{\epsilon}\sim\mathcal{N}(0,\mathbf{\textit{I}})}\left[
    \| \hat{\bm{\epsilon}}_{\theta}(\alpha_t \bm{x}_t+\sigma_t\bm{\epsilon},t,c,y) - \bm{\epsilon} \|_2^2
\right].
\end{equation}
Nonetheless, neither SDS nor VSD considers 3D consistency issues. Their optimization objectives may force NeRF to present a ``most-likely" diffusion-generated image in any camera view, resulting in the Janus problem.

\section{DreamControl}
\begin{figure*}[t]
    \centering
    \includegraphics[width=\textwidth]{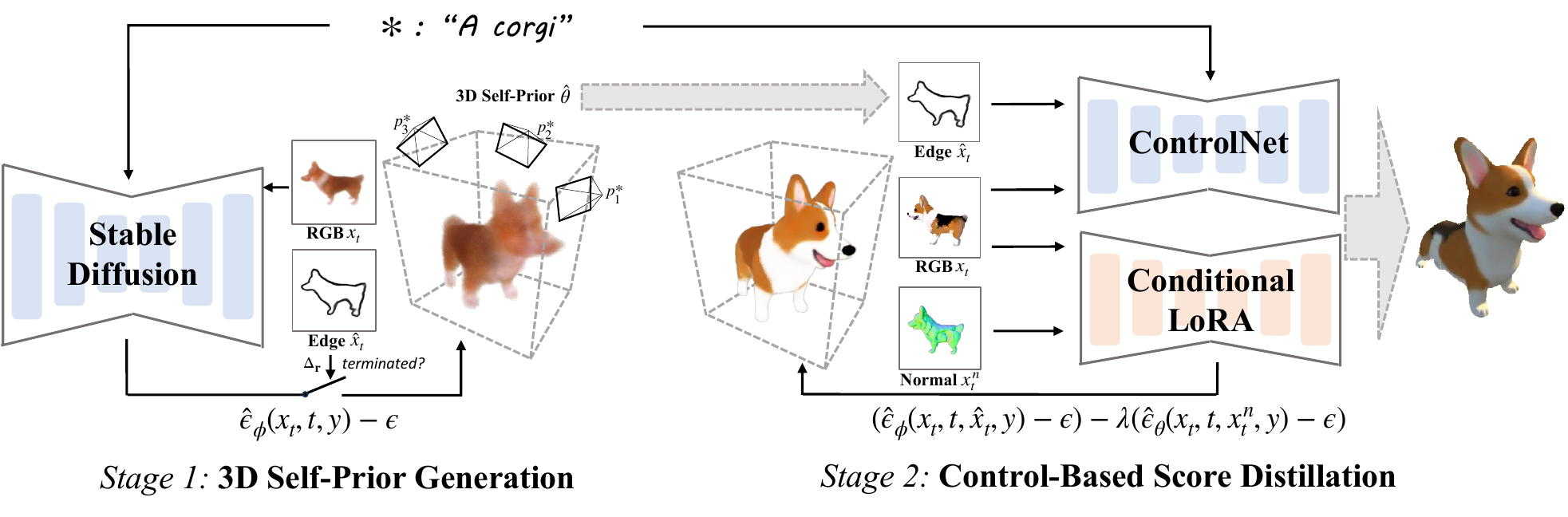}
    \caption{Overview of DreamControl. In the first stage, a coarse NeRF is optimized as a 3D self-prior $\hat{\theta}$, in which an adaptive viewpoint sampling $p^{\ast}$ and a boundary integrity metric $\Delta_{\mathbf{r}}$ are proposed to alleviate inconsistent generation. The prior $\hat{\theta}$ is then sent to the second stage as an input edge condition $\hat{x}_t$, in which a control-based score distillation can generate fine-grained textures and maintain geometries in the prior. A Conditional LoRA and a weighted score are further proposed to stabilize the optimization process.}
    \label{fig:method}
\end{figure*}

In this section, we introduce a two-stage 2D-lifting framework, DreamControl. As shown in Figure~\ref{fig:method}, we first generate a coarse NeRF with SDS, which is regarded as a 3D self-prior (Section~\ref{sec:nerf}). Then, we propose a control-based score distillation, generating high-quality texture while maintaining the geometry from previous priors (Section~\ref{sec:controlnet}).

\subsection{3D Self-Prior Generation}\label{sec:nerf}
In SDS optimization, the gradient direction is towards minimizing the distribution gap between rendered images in NeRF and generated images by 2D diffusion models,
\begin{equation}\label{eq:sds_target}
\mathop{\text{min}}\limits_{\theta \in \Theta}\mathcal{L}_{\text{SDS}}(\theta) \iff
\mathop{\text{min}}\limits_{\theta \in \Theta} D_{\text{KL}}(q^{\theta}_t(\bm{x}_t|y,c) || p_t(\bm{x}_t|y,c)),
\end{equation}
where $q^{\theta}_t$ and $p_t$ the distribution probability of the NeRF scene $\theta$ and a pre-trained 2D diffusion model, respectively.
However, there exists a viewpoint bias in the 2D model, which means it may not be able to acquire an accurate distribution $p_t(\bm{x}_t|y,c)$ for a given camera pose $c$.
When overfitting to the biased distribution $p_t$, optimization target in \cref{eq:sds_target} can degenerate to $D_{\text{KL}}(q^{\theta}_t(\bm{x}_t|y) || p_t(\bm{x}_t|y,\widetilde{c}))$, where $\widetilde{c}$ is the most frequent viewpoint. As a result, generated results may look similar in any camera view, giving rise to the Janus problem.

In this optimization process, two crucial aspects are the main cause of the problem, \textit{i.e.}, viewpoint bias and overfitted distribution. Accordingly, an adaptive viewpoint sampling and a boundary integrity metric are proposed in the following to alleviate potential inconsistent generation.

\noindent\textbf{Adaptive Viewpoint Sampling.} 2D diffusion models inherit viewpoint bias from the Internet training data, \textit{e.g.}, they tend to generate a person's front face rather than its back. To obtain a satisfied image, previous works~\cite{poole2022dreamfusion,armandpour2023re} attach view-dependent information to text prompts like ``\textit{front view}", but 2D networks cannot explicitly encode these view prompts. Some recent works~\cite{liu2023zero,zhao2023efficientdreamer,shi2023mvdream,liu2023syncdreamer,long2023wonder3d,li2023sweetdreamer} attempt to train a view-aware generative model with 3D data. However, texture fidelity and content generalizability are limited by available 3D training data.

Considering incorporating 2D diffusion models with 3D information is challenging, we present a solution with a new perspective, \ie, aligning 3D camera sampling with the viewpoint distribution of 2D diffusion.
Specifically, we modify the camera pose sampling $p(c)$ in SDS to fit the distribution in 2D. Given a text prompt ``$\ast$'', we make three view-dependent prompts, \textit{i.e.}, ``$\ast$, \textit{front view}'' ($y_1$), ``$\ast$, \textit{side view}'' ($y_2$), and ``$\ast$, \textit{back view}'' ($y_3$), and then take them to generate corresponding 2D images $\phi(y_i,t)$ with the diffusion model $\phi$ in time step $t$. 
Thus, the expected probability distribution of each view range $p^{\ast}$ can be calculated as,
\begin{equation}
    p^{\ast} = \text{softmax}([s_1,s_2,s_3]), s_i= \frac{1}{|T|} \sum_{t \in T} s_{\text{CLIP}}(y_i,\phi(y_i,t)),
\end{equation}
where $T$ is a set of timesteps. $s_{\text{CLIP}}$ denotes the CLIP similarity between text and image. In each view range, the viewpoint probability $p^{\ast}(c)$ follows a uniform distribution.
%
In this way, we have a NeRF representation $q^{\theta}_t(\bm{x}_t|y) = \int q^{\theta}_t(\bm{x}_t|y,c) p^{\ast}(c) dc$, which can get closer to $p_t(\bm{x}_t|y)$ in terms of camera pose $c$.

\noindent\textbf{Boundary Integrity Metric.} Albeit NeRF is modeled based on graphical principles, its reconstruction result highly depends on the quality of training data. Each ground-truth image and its own camera pose are both required, so that the prediction of color and density can be supervised by casting rays $\mathbf{r}$. However, it is difficult to instruct a 2D generative model to generate images that accurately match viewpoints. Even though our sampling strategy can alleviate the effects of viewpoint bias, 3D artifacts are still unavoidable in the overfitting circumstance. Previous works~\cite{poole2022dreamfusion,lin2023magic3d,metzer2023latent,Chen_2023_ICCV,wang2023prolificdreamer} hardly considered to check the occurrence of overfit automatically. In practice, they usually optimize 3D representations for a fixed number of iterations or manually terminate the optimization.

In this work, we propose a geometric terminated metric to avoid the possible overfit.
Specifically, we observe that NeRF generally can form a solid object without overfitting when the density between foreground and background starts to show a clear boundary.
Thus, we can detect the situation by calculating the difference between the density $\bm{\sigma}$ of all valid pixels and boundary pixels, \ie,
\begin{equation}
    \Delta_{\mathbf{r}} = \frac{1}{|\mathcal{R}_v|}\sum_{\mathbf{r} \in \mathcal{R}_v} \bm{\sigma}(\mathbf{r}) - \frac{1}{|\mathcal{R}_b|}\sum_{\mathbf{r} \in \mathcal{R}_b} \bm{\sigma}(\mathbf{r}),
\end{equation}
where $\mathcal{R}_v$ and $\mathcal{R}_b$ are the set of valid rays and boundary rays. 
And we terminate the optimization process when $\Delta_{\mathbf{r}}$ falls below our threshold $\delta_{\mathbf{r}}$.

With the adaptive viewpoint sampling and boundary integrity metric, a coarse shape $\hat{\theta}$ that keeps a reasonable geometry can be generated based on NeRF. We regard it as a 3D self-prior for the following control-based generation.

\subsection{Control-Based Score Distillation}\label{sec:controlnet}
The 3D prior $\hat{\theta}$ cannot capture fine-grained texture in a short optimization period. Previous multi-stage methods~\cite{lin2023magic3d,Chen_2023_ICCV,wang2023prolificdreamer} continually optimize the generation based on coarse shape, which contradicts our intention of early termination in Section~\ref{sec:nerf} and may still lead to overfitting. To generate high-quality texture while maintaining a reasonable geometry, we propose to treat  $\hat{\theta}$ as a conditional input and adopt ControlNet~\cite{zhang2023adding} as the optimization guidance.

Specifically, for one optimization step, we render an RGB image $\bm{x}$ from $\theta$ and a conditional image $\hat{\bm{x}}$ from $\hat{\theta}$ in the same camera pose. Then, ControlNet can supervise the generation by the predicted noise $\hat{\bm{\epsilon}}_{\phi}(\bm{x}_t,t,\hat{\bm{x}}_t,y)$ according to \cref{eq:sds}, which is demonstrated effective in previous works~\cite{shao2023control4d,yu2023hifi}.
However, these methods generally require high-quality conditions for the generation in ControlNet, \textit{e.g.}, depth maps or DSLR photos. Strong conditions such as depth may restrict the diversity of generated content. Meanwhile, our 3D prior $\hat{\theta}$ can only exhibit poor texture information, making it hard to extract photo-realistic images for generative control.

As for the issue of diversity, we take VSD~\cite{wang2023prolificdreamer} into consideration, which expands the generation range of a given prompt. Since photo-realistic images are not available from our 3D priors, we use the boundary mask from $\hat{\theta}$  as the condition $\hat{x}$, further weakening the control restriction. 
However, simply replacing the pre-trained diffusion model with ControlNet doesn't work well, as the pre-trained term $\hat{\bm{\epsilon}}_\phi$ and the LoRA term $\hat{\bm{\epsilon}}_\theta$ vary a lot, making it hard for the convergence of \cref{eq:lora}. To stabilize the optimization, we propose a conditional LoRA and a weighted score.

\noindent\textbf{Conditional LoRA.} In VSD, a LoRA is adopted to predict the noise of current NeRF scene $\hat{\bm{\epsilon}}_\theta(\bm{x}_t,t,c,y)$, in which a camera pose $c$ is additionally embedded. We are concerned that camera pose is a high-level semantic concept, which is hard to tokenize and imbibe through LoRA. The LoRA term $\hat{\bm{\epsilon}}_\theta$ may predict a noise unrelated to $c$, enlarging the gap with the pre-trained term $\hat{\bm{\epsilon}}_\phi$. Instead, we replace camera pose $c$ with the normal map $\bm{x}_t^n$ rendered by itself. Accordingly, a lightweight control LoRA is adopted to predict noise conditioned by $\bm{x}_t^n$. The training objective is similar to \cref{eq:lora}.

\noindent\textbf{Weighted Score.} The prediction of the LoRA term $\hat{\bm{\epsilon}}_\theta$ in the early stage lacks practical significance, as NeRF has not yet generated meaningful objects. To mitigate the possible disruption caused by LoRA at those early steps, we propose to incorporate a coefficient $\lambda$ into its loss term, which is changed along with training steps. Specifically, we rewrite $\hat{\bm{\epsilon}}_{\phi}(\bm{x}_t,t,y) - \hat{\bm{\epsilon}}_{\theta}(\bm{x}_t,t,c,y)$ in \cref{eq:vsd} as two terms,
\begin{equation}\label{eq:our_vsd}
    (\hat{\bm{\epsilon}}_{\phi}(\bm{x}_t,t,\hat{\bm{x}}_t,y)-\bm{\epsilon}) - \lambda (\hat{\bm{\epsilon}}_{\theta}(\bm{x}_t,t,\bm{x}_t^n,y)-\bm{\epsilon}).
\end{equation}
When $\lambda=0$, the loss function degenerates to SDS format, which can be regarded as a special case of VSD. We gradually increase $\lambda$ as the training step increases.

Following the training strategy of ProlificDreamer~\cite{wang2023prolificdreamer}, we alternately update the gradient of \cref{eq:vsd} and \cref{eq:lora}.

\subsection{Implementation Details}
We implement DreamControl based on threestudio~\cite{threestudio2023}. In the first stage, we adopt DeepFloyd~\cite{deepif} guidance in the SDS optimization. In the second stage, we use the scribble version of ControlNet~\cite{zhang2023adding} v1.1 as the pre-trained term and the stable diffusion~\cite{rombach2022high} v1.5 injected with ContraolLoRA~\cite{wu2023controllora} as the LoRA term. 3D prior is the rendered density mask from the NeRF result in the first stage, further processed by HEDdetector~\cite{xie15hed} implemented in ControlNet. The LoRA condition is the rendered normal map from the current NeRF. The classifier-free guidance scale (CFG) is set as 7.5 and 1.0 in pre-trained and LoRA terms, respectively. We increase the loss coefficient $\lambda$ from 0.5 to 0.75 linearly in the first 5,000 iterations. Please refer to the Suppl. for more details.
\section{Experiments}
In this section, extensive experiments are conducted to evaluate the generation quality of our proposed method DreamControl. We first show our text-to-3D generation results in Section~\ref{sec:text}, in which several state-of-the-art methods are compared in terms of geometry consistency and texture fidelity. In Section~\ref{sec:control}, we present controllable 3D generation tasks with our control-based optimization guidance, including user-guided generation and 3D animation. Finally, we conduct ablation studies in Section~\ref{sec:ablation}, demonstrating the effectiveness of our newly proposed designs.

\begin{table}[t]
    \centering
    \caption{Quantitative results. DreamControl surpasses the competing methods in all the evaluation metrics.}
    \begin{tabular}{l|ccc}
        \hline
        Method & JR(\%)$\downarrow$ & PS(\%)$\uparrow$ & CS(\%)$\uparrow$ \\
        \hline
        DreamFusion-IF~\cite{poole2022dreamfusion} & 36.67 & 10.01 & 26.36\\
        Magic3D~\cite{lin2023magic3d} & 53.33 & 16.13 & 26.59\\
        ProlificDreamer~\cite{wang2023prolificdreamer} & 56.67 & 20.81 & 26.69\\
        \hline
        Zero-1-to-3~\cite{liu2023zero} & 16.67 & 7.89 & 21.25\\
        MVDream~\cite{shi2023mvdream} & \textbf{10.00} & 17.70 & 26.17\\
        \hline
        DreamControl (ours) & \textbf{10.00} & \textbf{27.46} & \textbf{28.14}\\
        \hline
    \end{tabular}
    \label{tab:exp}
    \vspace{-0.5em}
\end{table}

\begin{figure*}[t]
    \centering
    \includegraphics[width=\textwidth]{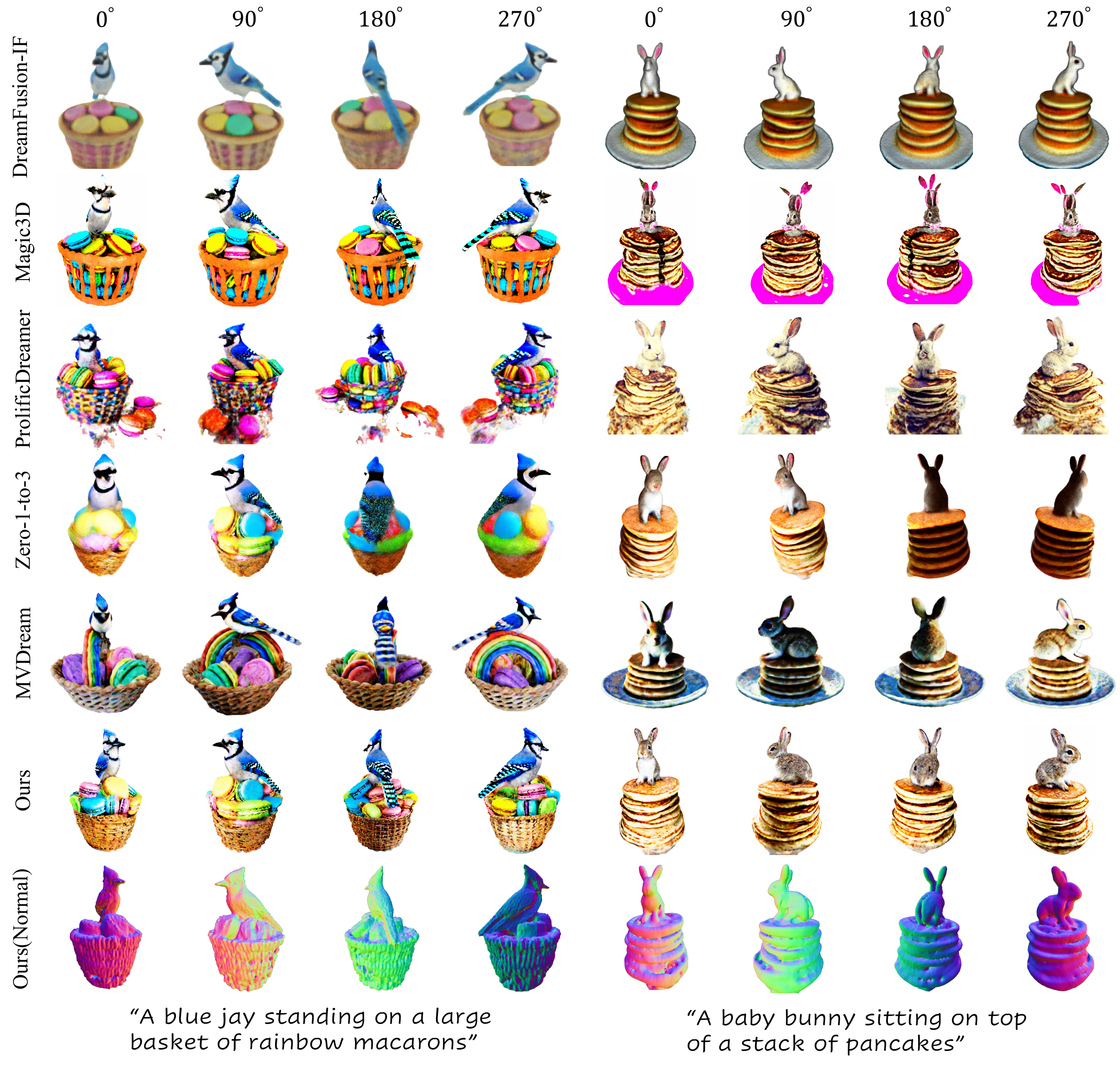}
    \caption{Qualitative results. Compared with other methods, DreamControl enjoys high-consistency geometry and high-fidelity texture.}
    \label{fig:exp1}
\end{figure*}

\subsection{Text-to-3D Generation}\label{sec:text}
We compare DreamControl with five 3D generation methods: (1) DreamFusion~\cite{poole2022dreamfusion}, the early work in 2D-lifting methods, (2) Magic3D~\cite{lin2023magic3d}, the first two-stage optimization method, (3) ProlificDreamer~\cite{wang2023prolificdreamer}, a high-fidelity optimization method, (4) Zero-1-to-3~\cite{liu2023zero}, a view-conditional diffusion model, and (5) MVDream~\cite{shi2023mvdream}, a multi-view diffusion model. Since most of these methods haven't released official implementation, we use the reproduction in threestudio~\cite{threestudio2023}. Note that the reproduction can be different from the original implementation. For example, DreamFusion adopts an unreleased diffusion model Imagen~\cite{saharia2022photorealistic}, which is replaced with DeepFloyd~\cite{deepif} in our comparison. We provide quantitative and qualitative results in the following.

\begin{figure}[t]
    \centering
    \includegraphics[width=0.45\textwidth]{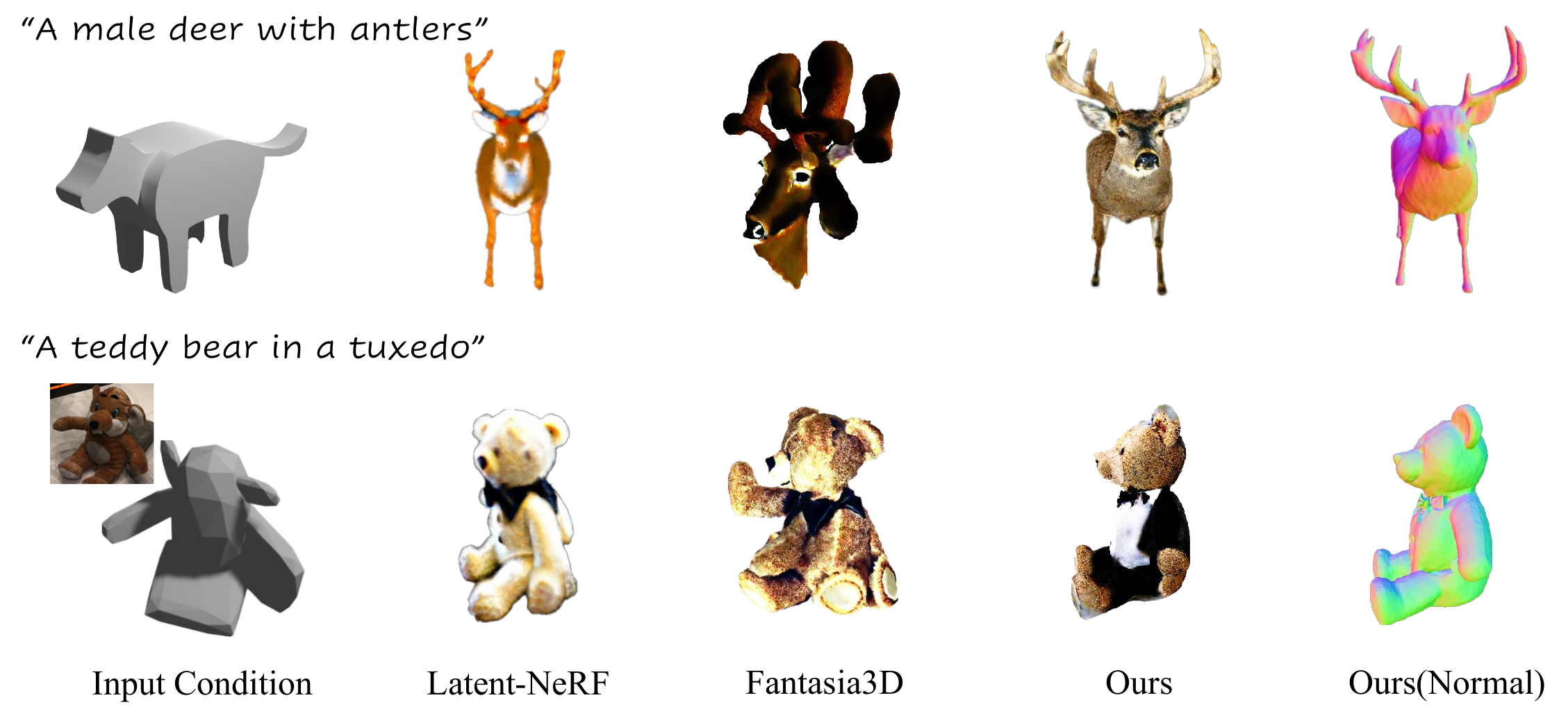}
    \caption{User-guided generation. DreamControl is flexible to loose input conditions, generating fine-grained content with a 3D sketch or even a coarse layout.}
    \label{fig:exp2}
    \vspace{-0.5em}
\end{figure}

\begin{figure}[t]
    \centering
    \includegraphics[width=0.45\textwidth]{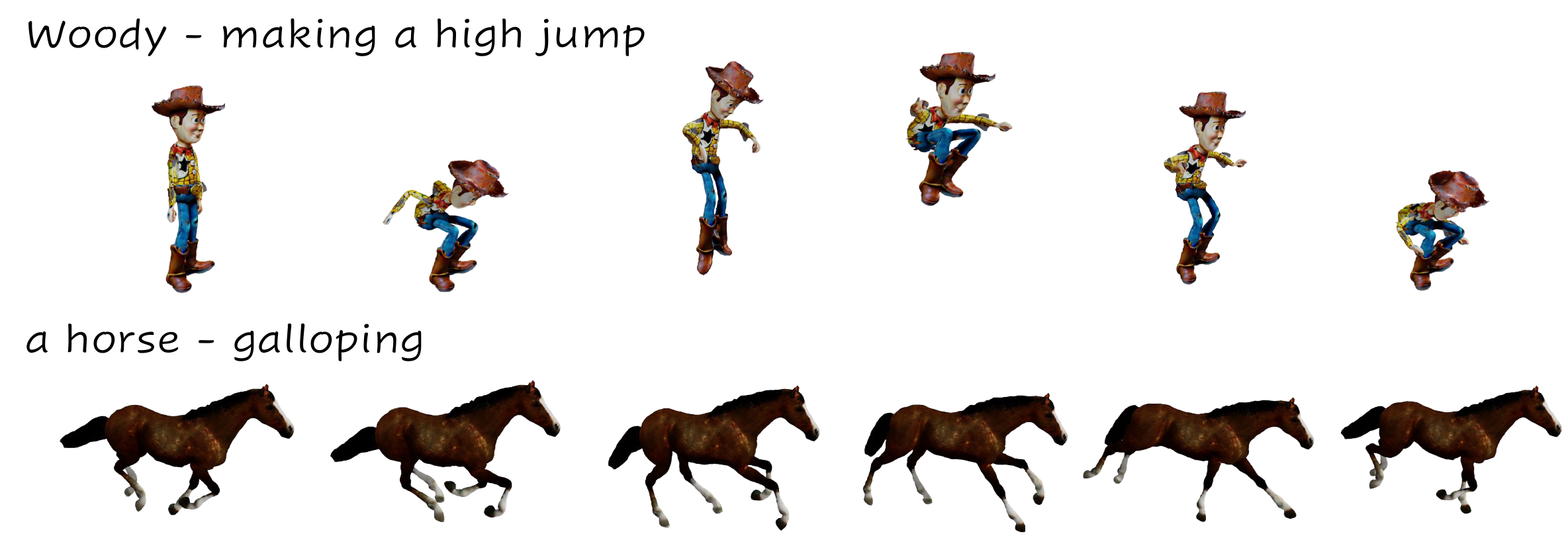}
    \caption{3D Animation. DreamControl can generate 3D content conditioned by a template skeleton, which is naturally bound with the skeleton after generation.}
    \label{fig:exp3}
    \vspace{-1.5em}
\end{figure}

\noindent\textbf{Quantitative Results.}
We select 30 text prompts from the galleries of DreamFusion, Magic3D, and ProlificDreamer for quantitative experiments. To evaluate the consistency of 3D geometries, we count the number of inconsistent 3D content generated in each method, regarded as the occurrence rate of the Janus problem (JR). To evaluate the fidelity of textures, we render multi-view images and introduce PickScore (PS)~\cite{kirstain2023pick} for comparison, which is an aesthetic metric used to measure the quality of 2D content. Moreover, we adopt CLIP-Score (CS), verifying the text consistency of generation. See the Suppl. for the details of evaluation metrics. As shown in Table~\ref{tab:exp}, our DreamControl outperforms other methods in all the metrics. The first three methods face a serious Janus problem. JR grows as PS increases, which indicates that the overfitting of optimization could exacerbate 3D inconsistency. Zero-1-to-3 and MVDream incorporate 3D prior knowledge into 2D diffusion, capable of generating high-consistent geometries. However, they suffer from low-quality texture and text-irrelevant generation, according to PickScore and CLIP-Score. We have analyzed that 3D training data tend to exhibit a cartoonish style and lack generalizability due to the limited scale. The results further demonstrate this point. In comparison, DreamControl adopts a coarse NeRF as a 3D self-prior, enjoying high-quality 3D generation in both geometry and texture, as well as the consistency of text input.

\noindent\textbf{Qualitative Results.}
To compare the generation results qualitatively, we select two classic text prompts from quantitative experiments and show the corresponding multi-view rendered images in Figure~\ref{fig:exp1}. Textures in DreamFusion-IF are over-smoothed. The rabbit's ears are still red in the back view, implying the risk of the multi-face issue. Yet DreamFusion basically generates reasonable geometries, demonstrating our assumption that NeRF can be used as a self-generated 3D prior. The two-stage method Magic3D presents more detailed textures but 3D inconsistent problems emerge, \textit{e.g.}, two beaks in the generated jay and three ears in the generated rabbit. ProlificDreamer can generate high-fidelity textures, while the Janus problem still exists. In particular, the other face is generated in the back view of the rabbit. Zero-1-to-3 and MVDream create highly consistent geometries. However, the generated textures are quite rough, \textit{e.g.}, the macarons and pancakes. Moreover, MVDream generates an unrelated rainbow in the first prompt, which means 3D prior is not general enough to handle detailed text descriptions. In contrast, our DreamControl can generate 3D content with high-consistency geometries and high-fidelity textures. The visualization is consistent with quantitative results.

\subsection{Controllable 3D Generation}\label{sec:control}
Thanks to our control-based optimization guidance, DreamControl can also be applied to controllable 3D generation tasks. We present user-guided generation in Figure~\ref{fig:exp2} and 3D animation in Figure~\ref{fig:exp3}.

\noindent\textbf{User-Guided Generation.}
User-guided generation is to create 3D content based on a text prompt, as well as a spatial sketch. Compared with previous works Latent-NeRF~\cite{metzer2023latent} and Fantasia3D~\cite{Chen_2023_ICCV}, our method is flexible to looser input conditions. In the first row of Figure~\ref{fig:exp2}, we use a simple animal shape as a template, successfully generating a male deer with antlers. Latent-NeRF adopts a sketch loss to minimize the geometric distance between generation and the input condition. Compared with it, our result is much more detailed in textures. Fantasia3D initializes the 3D representation with input conditions and then gradually refines an expected geometry. However, it fails to attach antlers to the animal body, generating a weird deer head instead. In the second row, we present the generation task based on a harder condition, \textit{i.e.}, multi-view images. We use a classic multi-view image set in DTU MVS dataset~\cite{jensen2014large} and reconstruct a coarse layout with~\cite{monnier2023differentiable}. Conditioned on that coarse shape, our method can generate a teddy bear in a fine-grained tuxedo, while the other two create teddy bears with multiple faces and legs. The results exhibit the effectiveness of our guidance on flexible conditions.

\noindent\textbf{3D Animation.}
To animate a 3D object, the common practice is to bind the object with a template skeleton, also known as rigging, which is a time-consuming work that demands related technical expertise. As a result, 3D animation is an expensive production currently. Instead, our method can generate 3D objects conditioned by a given skeleton, which can be directly bound with the skeleton after generation. Different from previous works TADA~\cite{liao2023tada} that is based on a human-template SMPL-X~\cite{pavlakos2019expressive}, DreamControl can adapt to all kinds of templates like animal, machine, \textit{etc}. As shown in Figure~\ref{fig:exp3}, we can easily animate the generated Woody and horse.

\subsection{Ablation Studies}\label{sec:ablation}
\begin{figure}[t]
    \centering
    \includegraphics[width=0.47\textwidth]{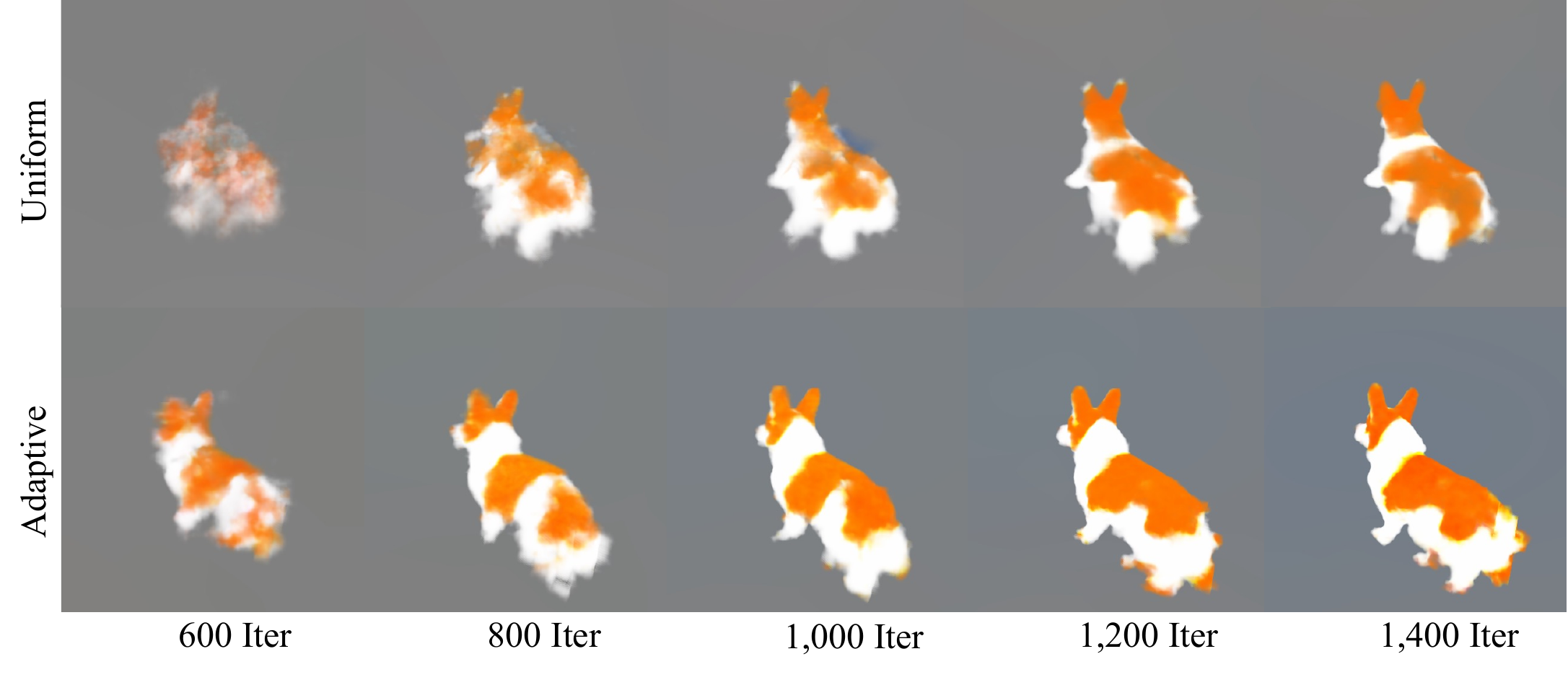}
    \vspace{-1em}
    \caption{Ablation study on viewpoint sampling and termination metric. Our optimization can avoid the generation of extra legs.}
    \label{fig:ab1}
    \vspace{-0.5em}
\end{figure}

\begin{figure}[t]
    \centering
    \includegraphics[width=0.47\textwidth]{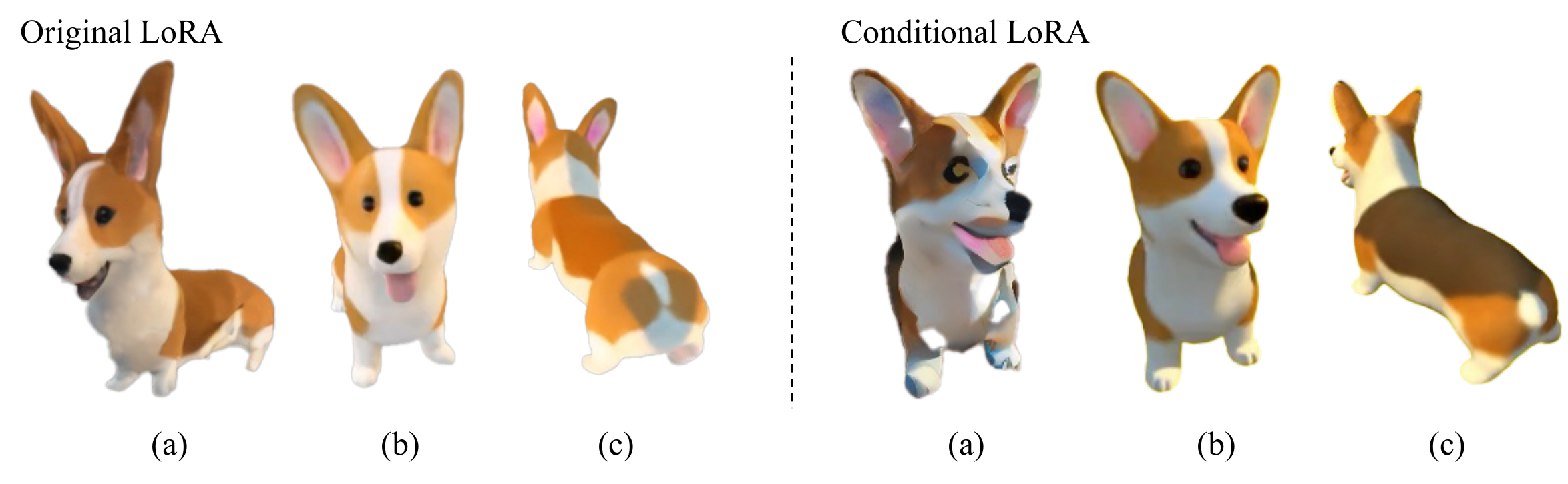}
    \vspace{-1em}
    \caption{Ablation study on conditional LoRA. (a) is the LoRA sampling result in the front view. (b) and (c) are 3D generation results in the front view and the back view. Our conditional LoRA can precisely present the current representation from a correct viewpoint, improving the generation quality.}
    \label{fig:ab2}
    \vspace{-0.5em}
\end{figure}

\begin{figure}[t]
    \centering
    \includegraphics[width=0.47\textwidth]{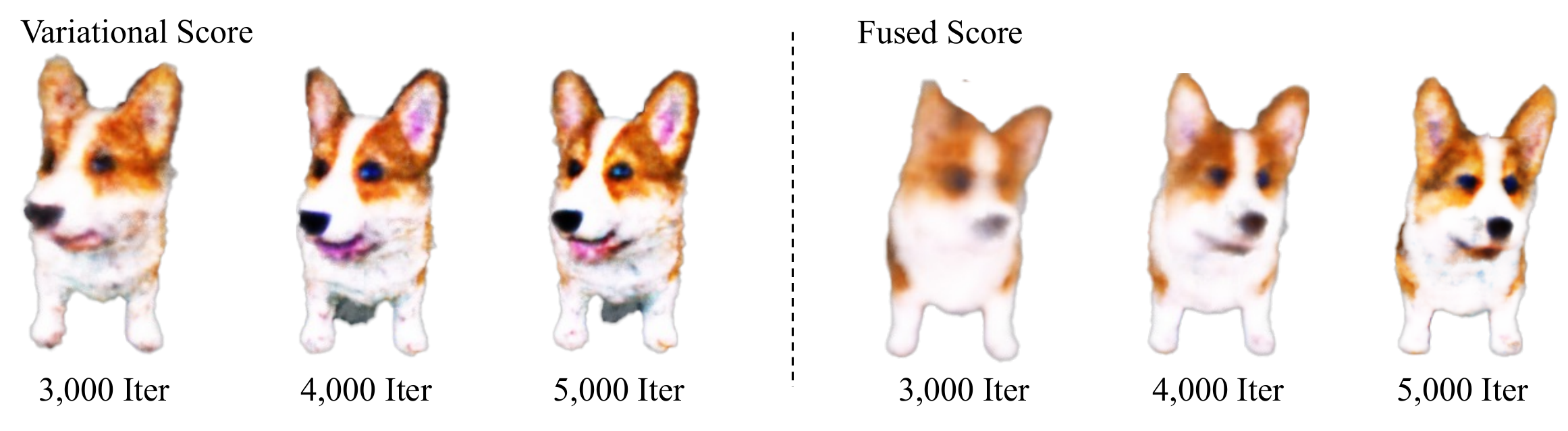}
    \caption{Ablation study on fused score distillation. Compared with VSD loss, our loss stabilizes the generation in the early stage.}
    \label{fig:ab3}
    \vspace{-1em}
\end{figure}

To further verify the effectiveness of our components, we take the generation of a corgi as an example, comparing results with or without our designs in Figures~\ref{fig:ab1},~\ref{fig:ab2}, and ~\ref{fig:ab3}.

\noindent\textbf{Viewpoint Sampling and Optimization Termination.} We compare our adaptive viewpoint sampling with uniform sampling in Figure~\ref{fig:ab1}. In the uniform sampling, the corgi's extra legs grow up evenly with normal legs, making it hard to find an appropriate timestep for terminating the optimization. Differently, the extra legs do not appear until 1,000 iterations with our sampling strategy. However, the multi-leg issue still exists, indicating that termination is necessary.

\noindent\textbf{Conditional LoRA.} We compare our conditional LoRA with ProlificDreamer's original LoRA in Figure~\ref{fig:ab2}. (a) is the sampling result of LoRA in the front view. With a camera pose $c$, the original LoRA can hardly generate 2D content in an accurate viewpoint. In contrast, our conditional LoRA successfully learns the current scene with a normal map $\mathbf{n}$. Due to the estimation bias of the original LoRA, the generated corgi is incompatible with ours. Especially in the back view, its ears are colored pink, and its body and bottom are disproportionately scaled.

\noindent\textbf{Fused Score.} We compare our fused score with the variational score of VSD in Figure~\ref{fig:ab3}. VSD optimization formulates a clear outline of a corgi after 3,000 iterations, which seems a stronger supervision than ours. However, the corgi contains too much noise, especially near the generation boundary. By restricting LoRA in the early stage, our optimization is much more stable than VSD.
\section{Conclusion}
\begin{figure}[t]
    \centering
    \includegraphics[width=0.47\textwidth]{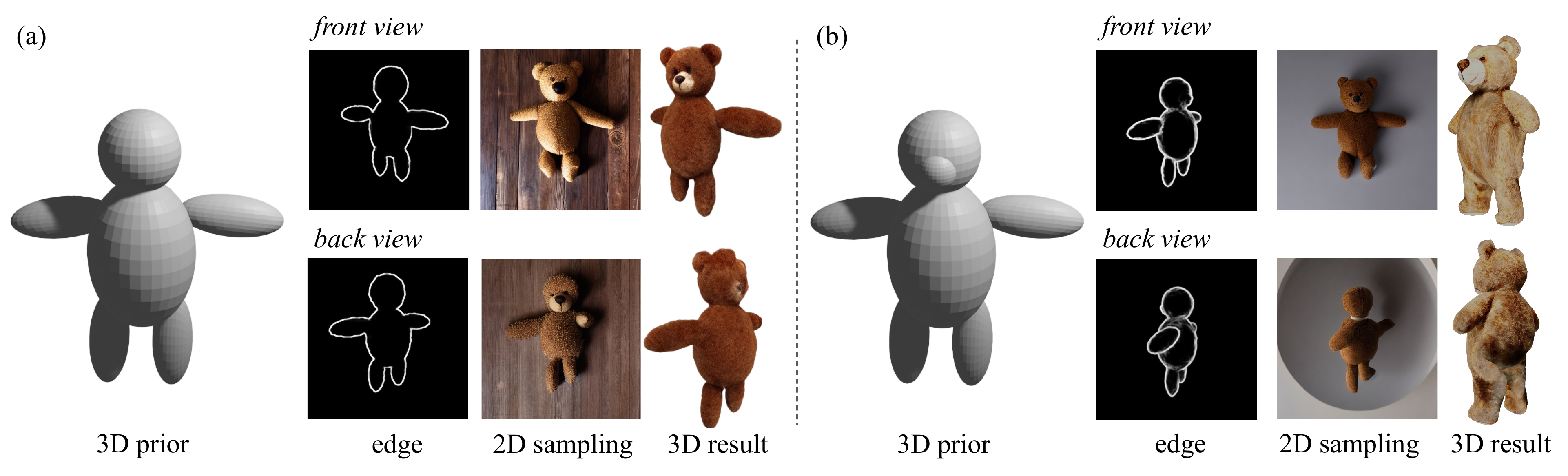}
    \caption{Failure case. (a) When edge conditions from different views are similar, ControlNet may fail to provide correct guidance. (b) A possible solution is to add more details in one side.}
    \label{fig:failure}
    \vspace{-1.5em}
\end{figure}

In this work, we present a two-stage framework DreamControl to improve the geometry consistency of 3D generation. By optimizing coarse NeRF, our method is able to obtain 3D self-prior for the following generation. A control-based score distillation is further proposed, generating fine-grained texture while maintaining the prior geometry. DreamControl can create high-quality 3D content with high-consistency geometries and high-fidelity textures. The control-based guidance can also be applied to controllable tasks including user-guided generation and 3D animation.

\noindent\textbf{Limitation.}
Although DreamControl maintains the geometry consistency with prior-conditional guidance, it may fail in some extreme cases where priors look similar in different views. For example, the left prior in Figure~\ref{fig:failure} looks exactly the same in opposite views, and ControlNet thus provides incorrect guidance in back views, leading to multi-face generation. Fortunately, this problem is avoidable by enlarging the view differences, \textit{e.g.}, adding more details to the front view of 3D priors like the right prior in Figure~\ref{fig:failure}.

\noindent\textbf{Broader Impact.}
DreamControl allows automatic text-to-3D generation and some more personalized 3D generation, reducing the cost of 3D content creation. However, like most of the other generative models, a risk of generating malicious content exists.

\section*{Acknowledgements}
This work was supported by National Key RD Program of China under Grant No. 2021ZD0112100, and the National Natural Science Foundation of China (NSFC) under Grant No. U19A2073.

{
    \small
    \bibliographystyle{ieeenat_fullname}
    \bibliography{main}
}

\clearpage
\setcounter{page}{1}
\maketitlesupplementary

In this supplementary, we provide more information on our implementation details (Section~\ref{sec:imple_detail}) and evaluation metrics (Section~\ref{sec:metric}). To demonstrate the effectiveness of our control-based guidance on maintaining 3D prior, we provide a comparison with a two-stage ProlificDreamer (Section~\ref{sec:2stage}). For more visualization results in the format of $360^\circ$, please refer to the attached HTML file in \textit{results} folder.

\section{Implementation Details}\label{sec:imple_detail}
\begin{figure*}[t]
    \centering
    \includegraphics[width=\textwidth]{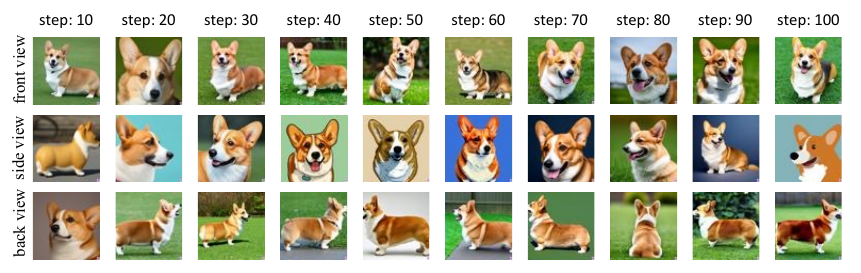}
    \caption{Visualization of generation in different view-dependent prompts. Generated corgis in the front view present a higher confidence than the other two views.}
    \label{fig:viewpoint}
\end{figure*}

\noindent\textbf{Adaptive Viewpoint Sampling.}
We use a standard DreamFusion-IF framework in the first stage, in which the uniform viewpoint sampling is replaced with an adaptive sampling. To model an adaptive distribution, we evaluate the generation confidence in different views. The rotation angle ranges of the front view, the side view, and the back view are $[-60^\circ, 60^\circ)$, $[-120^\circ, -60^\circ)\cup[60^\circ, 120^\circ)$, $[-180^\circ, -120^\circ)\cup[120^\circ, 180^\circ)$, respectively. For each view, we denoise the latent code with a set of timesteps. Here, we set the timestep set $T$ as $\{10,20,30,40,50,60,70,80,90,100\}$. We calculate the average CLIP similarity of all the timesteps, and softmax similarity scores of the three views as the view confidence. During optimization, the distribution of viewpoint sampling is proportional to confidence. Take generating corgi as an example, as shown in Figure~\ref{fig:viewpoint}, we denoise 3 views with 10 different timesteps. The generation confidences of front-view, side-view, and back-view are 68.75\%, 4.75\%, 26.51\%. Consequently, the final probabilities of viewpoint distribution are $p^\ast_1=68.75\%$, $p^\ast_2=4.75\%$, and $p^\ast_3=26.51\%$.

\begin{figure*}[t]
    \centering
    \includegraphics[width=\textwidth]{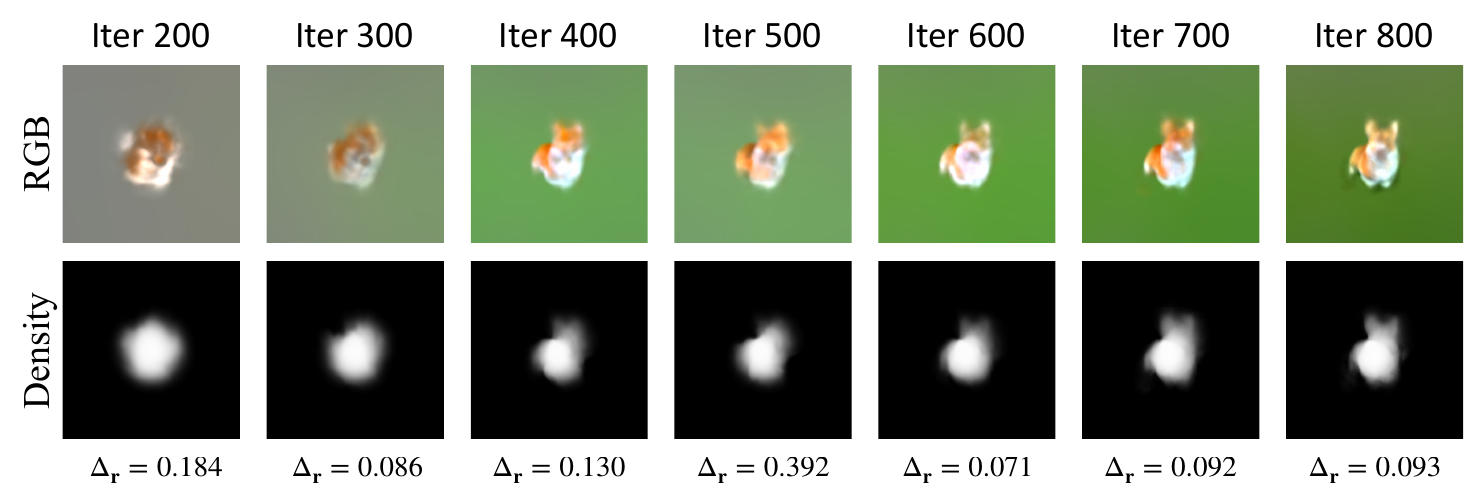}
    \caption{Visualization of the change in density difference $\Delta_{\mathbf{r}}$. As $\Delta_{\mathbf{r}}$ decreases, the NeRF representation gradually forms a solid geometry.}
    \label{fig:metric}
\end{figure*}

\noindent\textbf{Boundary Integrity Metric.}
Our terminated condition in the first stage is related to the density difference between all valid rays and edge rays in NeRF scenes. The edge is detected by HEDdetector~\cite{xie15hed} implemented in ControlNet. We calculate the average density of each ray set and terminate the optimization when $\Delta_{\mathbf{r}}$ is less than 0.1 for three consecutive checkpoints. We set a checkpoint for every 100 iterations. We also provide an example of generating corgi in Figure~\ref{fig:metric}. As the density difference decreases, the NeRF representation gradually forms a solid geometry.

\noindent\textbf{Mesh Extraction.}
Our two-stage optimization can achieve promising visual results under the representation of NeRF, while the surfaces may exhibit severe irregularities if we export NeRF results to mesh objects for applications like animation. To obtain higher-quality 3D models for broader applications, we further refine the mesh representation based on DMTet~\cite{shen2021deep}.
We employ the same ControlNet guidance under NeRF representation to optimize the refinement of SDF representation. Specifically, we convert NeRF into a DMTet object and retain its texture information. We initially optimize its geometry to achieve smooth and regular surfaces. Subsequently, we optimized its texture to remove the noise generated during the geometry optimization. Note that, the optimization guidance in the SDF phase is exactly the same with the NeRF phase. The parameters of conditional LoRA on the NeRF training are loaded before the SDF optimization, which is frozen during the geometry phase and unfrozen during the texture phase.

\noindent\textbf{Inference Time.}
In the the first stage, as the optimization is terminated automatically based on our metric, it usually takes less than 10 minutes for the generation of 3D self-prior. In the second stage, we optimize the generation for 15,000 iterations, which spend around 2 hours for a given prompt input. The total inference time is basically equivalent to previous text-to-3D methods like Fantasia3D~\cite{Chen_2023_ICCV} and ProlificDreamer~\cite{wang2023prolificdreamer}. For further extracting the mesh representation, since the color information from the NeRF phase is already of high quality, in the SDF phase, we only need to optimize the geometry and suppress texture noise appropriately. As a result, the entire process of mesh extraction is relatively fast, taking around 30 minutes.

\noindent\textbf{Pseudocode.}
We provide pseudocode in Algorithm~\ref{alg:nerf},\ref{alg:sdf} to summarize the NeRF phase and the SDF phase in our framework, respectively.

\begin{algorithm}[t] 
    \centering 
    \small
    \caption{DreamControl - NeRF Phase}
    \label{alg:nerf} 
    \begin{algorithmic}[1] 
        \STATE\textbf{Input}: text prompt y
        \STATE\textbf{Load}: stable diffusion v1.5 $\phi_{\text{sd}}$ and DeepFloyd XL-v1.0 $\phi_{\text{if}}$
        \STATE $\blacktriangleright$ \textbf{Preprocess: Viewpoint Confidence Analysis}
        \STATE $y_{1,2,3}=y+\text{``, front/side/back view"}$
        \FOR{$t=10,20,\dots,100$}
            \STATE Generate images $\bm{x}_{y_1}^t$, $\bm{x}_{y_2}^t$, $\bm{x}_{y_3}^t$ with $\phi_{\text{if}}$ in the timestep $t$
        \ENDFOR
        \STATE $p^\ast_{1,2,3}=\text{softmax}(s_\text{CLIP}(y_{1,2,3},[\{\bm{x}_{y_1}^t\},\{\bm{x}_{y_2}^t\},\{\bm{x}_{y_3}^t\}]))$
        \STATE $\blacktriangleright$ \textbf{Stage 1: 3D Self-Prior Generation}
        \STATE \textbf{initialize} a NeRF scene $\theta$
        \WHILE{\text{Not} $\Delta_{\mathbf{r}}<0.1$ for three consecutive checkpoints}
            \STATE Sample a camera $c$ based on $p^\ast$ and a timestep $t$
            \STATE Render $\theta$ at pose $c$, for RGB image $\bm{x}=\mathbf{g}(\theta,c)$
            \STATE $\theta \gets \theta - \mathbb{E}\left[\omega(t)\left( \hat{\bm{\epsilon}}_{\phi_{\text{if}}}(\bm{x}_t,t,y) - \bm{\epsilon} \right)\frac{\partial \bm{x}}{\partial \theta}\right]$
        \ENDWHILE
        \STATE 3D self-prior $\hat{\theta} = \theta$
        \STATE $\blacktriangleright$ \textbf{Stage 2: Control-Based Score Distillation}
        \STATE \textbf{initialize} a NeRF scene $\theta$ and a conditional LoRA $\phi_\theta$
        \WHILE{not converged}
            \STATE Sample a camera $c$ and a timestep $t$
            \STATE Render $\theta$ for RGB image $\bm{x}$ and normal map $\bm{x}^n$ at pose $c$
            \STATE Render $\hat{\theta}$ and detect an edge mask $\hat{\bm{x}}$ at pose $c$
            \STATE Update $\lambda$, weighted score $\mathcal{L}_\text{score}(\bm{x}_t,\hat{\bm{x}}_t,\bm{x}_t^n,t)=(\hat{\bm{\epsilon}}_{\phi_{\text{sd}}}(\bm{x}_t,t,\hat{\bm{x}}_t,y)-\bm{\epsilon}) - \lambda (\hat{\bm{\epsilon}}_{\theta}(\bm{x}_t,t,\bm{x}_t^n,y)-\bm{\epsilon})$
            \STATE $\theta \gets \theta - \mathbb{E}\left[\omega(t) \mathcal{L}_\text{score}(\bm{x}_t,\hat{\bm{x}}_t,\bm{x}_t^n,t)\frac{\partial \bm{x}}{\partial \theta}\right]$
            \STATE $\phi_\theta \gets \phi_\theta - \nabla_{\phi_\theta}\mathbb{E}\left[\| \hat{\bm{\epsilon}}_{\theta}(\alpha_t \bm{x}_t+\sigma_t\bm{\epsilon},t,c,y) - \bm{\epsilon} \|_2^2\right].$
        \ENDWHILE
        \RETURN the optimized NeRF representation $\theta$
    \end{algorithmic}
\end{algorithm}

\begin{algorithm}[t] 
    \centering 
    \small
    \caption{DreamControl - SDF Phase}
    \label{alg:sdf} 
    \begin{algorithmic}[1] 
        \STATE\textbf{Input}: text prompt y
        \STATE\textbf{Load}: $\phi_{\text{sd}}$, $\phi_\theta$, and $\theta$
        \STATE $\blacktriangleright$ \textbf{Stage 1: Geometry Refinement}
        \STATE \textbf{initialize} a DMTet mesh from $\theta$, parameterized by $m$
        \WHILE{not converged}
            \STATE Sample a camera $c$ and a timestep $t$
            \STATE Render $m$ for RGB image $\bm{x}$ and normal map $\bm{x}^n$ at pose $c$
            \STATE Render $\hat{\theta}$ and detect an edge mask $\hat{\bm{x}}$ at pose $c$
            \STATE $m \gets m - \mathbb{E}\left[\omega(t)\left( \hat{\bm{\epsilon}}_{\phi_{sd}}(\bm{x}_t,t,y) - \hat{\bm{\epsilon}}_{\theta}(\bm{x}_t,t,\bm{x}_t^n,y) \right)\frac{\partial \bm{x}}{\partial m}\right]$
        \ENDWHILE
        \STATE $\blacktriangleright$ \textbf{Stage 2: Texture Refinement}
        \WHILE{not converged}
            \STATE Sample a camera $c$ and a timestep $t$
            \STATE Render $m$ for RGB image $\bm{x}$ and normal map $\bm{x}^n$ at pose $c$
            \STATE Render $\hat{\theta}$ and detect an edge mask $\hat{\bm{x}}$ at pose $c$
            \STATE $m \gets m - \mathbb{E}\left[\omega(t)\left( \hat{\bm{\epsilon}}_{\phi_{sd}}(\bm{x}_t,t,y) - \hat{\bm{\epsilon}}_{\theta}(\bm{x}_t,t,\bm{x}_t^n,y) \right)\frac{\partial \bm{x}}{\partial m}\right]$
            \STATE $\phi_\theta \gets \phi_\theta - \nabla_{\phi_\theta}\mathbb{E}\left[\| \hat{\bm{\epsilon}}_{\theta}(\alpha_t \bm{x}_t+\sigma_t\bm{\epsilon},t,c,y) - \bm{\epsilon} \|_2^2\right].$
        \ENDWHILE
        \RETURN the optimized SDF representation $m$
    \end{algorithmic}
\end{algorithm}

\section{Evaluation Metrics}\label{sec:metric}

\begin{figure}[t]
    \centering
    \includegraphics[width=0.47\textwidth]{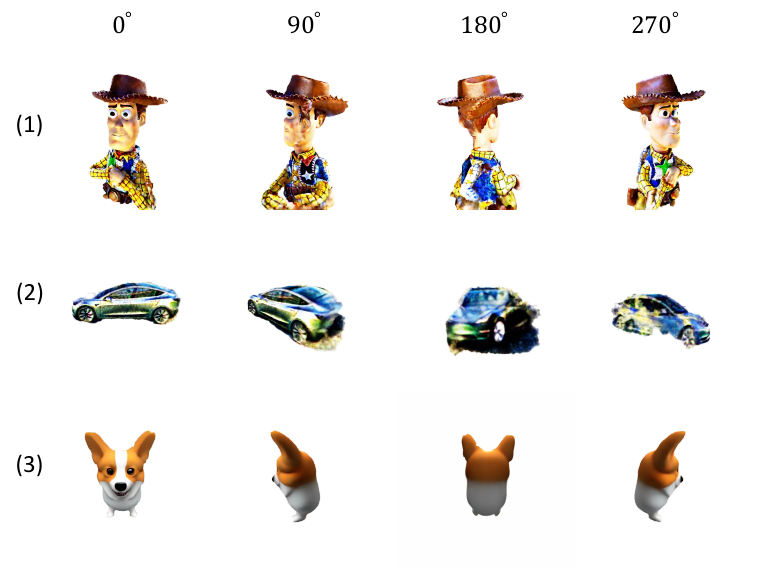}
    \caption{Visualization of Janus problems. (1) multi-face, multi-hand, multi-leg, or similar issues; (2) obvious content drift; (3) serious paper-thin generation.}
    \label{fig:janus}
\end{figure}

\noindent\textbf{Janus Rate.}
To evaluate the geometry consistency, we count the occurrence rate of the Janus problem (JR). As shown in Figure~\ref{fig:janus}, we count generation as a Janus problem based on the following situations: (1) multi-face, multi-hand, multi-leg, or similar issues; (2) obvious content drift; (3) serious paper-thin generation. We calculate JR as the number of inconsistent content out of the number of all generated objects.

\noindent\textbf{Pick-Score.}
Pick-Score (PS)~\cite{kirstain2023pick} is a CLIP-based scoring model, which is trained with Pick-a-Pic, a large, open dataset of text-to-image prompts and real users’ preferences over generated images. As a result, it can exhibit superhuman performance on the task of predicting human preferences. In 3D evaluation, we compare multi-view rendered images of generated objects to measure the preference. For a given text prompt $y$, we render N-view images $\{x_{y,i}^k\}_{i=1}^N$ from the generated object of method $k$. The pick score $\text{ps}$ for $y$ is formulated as,
\begin{equation}
    \text{ps}(y) = \frac{1}{N} \sum_{i=1}^N s_\text{pick}(y, [\bm{x}_{y,i}^1, \bm{x}_{y,i}^2, \dots, \bm{x}_{y,i}^k]),
\end{equation}
where $s_\text{pick}$ is the CLIP-based scoring function. For each method $k$, the final pick score is calculated as the average score of all the text prompts, \textit{i.e.}, $\frac{1}{|Y|} \sum_{y\in Y} \text{ps}(y,k)$.

\noindent\textbf{CLIP-Score.}
CLIP-Score (CS) is based on the CLIP~\cite{radford2021learning} similarity. For each prompt $y$, we render one image from the corresponding generated object of method $k$, \textit{$\bm{x}^k_y$}. The rendering viewpoint is a fixed camera pose at a $45^\circ$ angle of elevation and a $30^\circ$ angle of rotation. The final score is calculated as the average score of all the text prompts, \textit{i.e.}, $\frac{1}{|Y|} \sum_{y\in Y} s_\text{CLIP}(y,\bm{x}^k_y)$. Note that, differently from Pick-Score, we don't evaluate the CLIP-score on multi-view images because the Janus problem could ironically increase the CLIP similarity.

\section{Two-Stage Comparison}\label{sec:2stage}
To further demonstrate that our control-based guidance can maintain the 3D self-prior obtained in the first stage, we compare our second stage with two-stage ProlificDreamer~\cite{wang2023prolificdreamer}. For the two-stage ProlificDreamer, we use the 3D self-prior in our first stage for initialization.

\begin{figure}[t]
    \centering
    \includegraphics[width=0.45\textwidth]{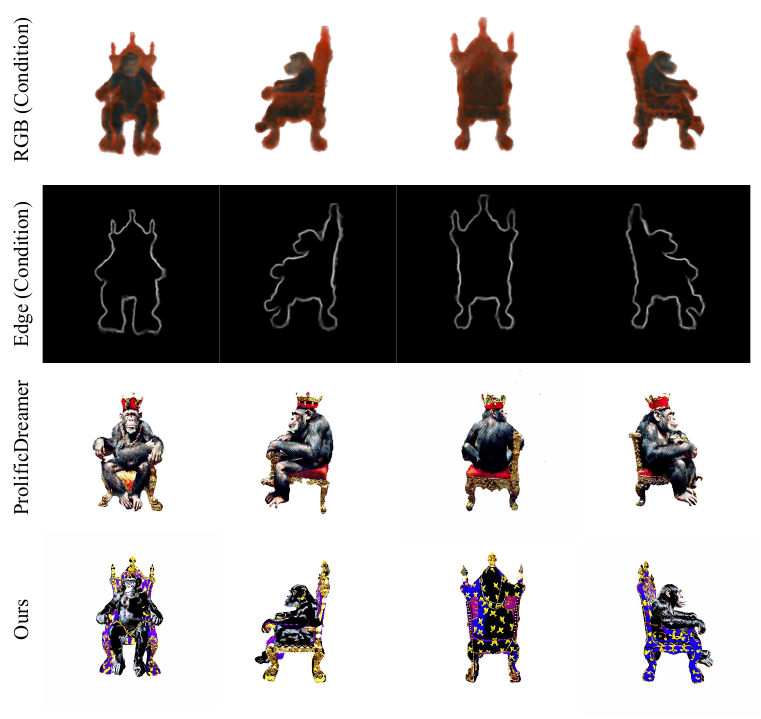}
    \caption{Visualization of two-stage generation comparison.}
    \label{fig:2stage}
\end{figure}

As shown in Figure~\ref{fig:2stage}, we take the prompt ``A chimpanzee dressed like Henry VIII king of England" as an example. The final generated chimpanzee of ProlificDreamer is quite different from its initialization, presenting multiple faces and multiple hands. Although DreamControl only uses the edge condition, our result can successfully maintain the geometry of the input condition. The results demonstrate that continually optimizing a coarse shape may still lead to overfitting issues.

\section{More Visualization Results}
Please refer to the attached HTML file in \textit{results} folder for more visualization results in the format of $360^\circ$ video.


\end{document}